# DCEdit: Dual-Level Controlled Image Editing via Precisely Localized Semantics

Yihan Hu[1,2*]  Jianing Peng[1]  Yiheng Lin[1,2]  Ting Liu[2]  Xiaochao Qu[2]  Luoqi Liu[2]

Yao Zhao[1]  Yunchao Wei[1†]

[1]Institute of Information Science, Beijing Jiaotong University  [2]MT Lab, Meitu Inc

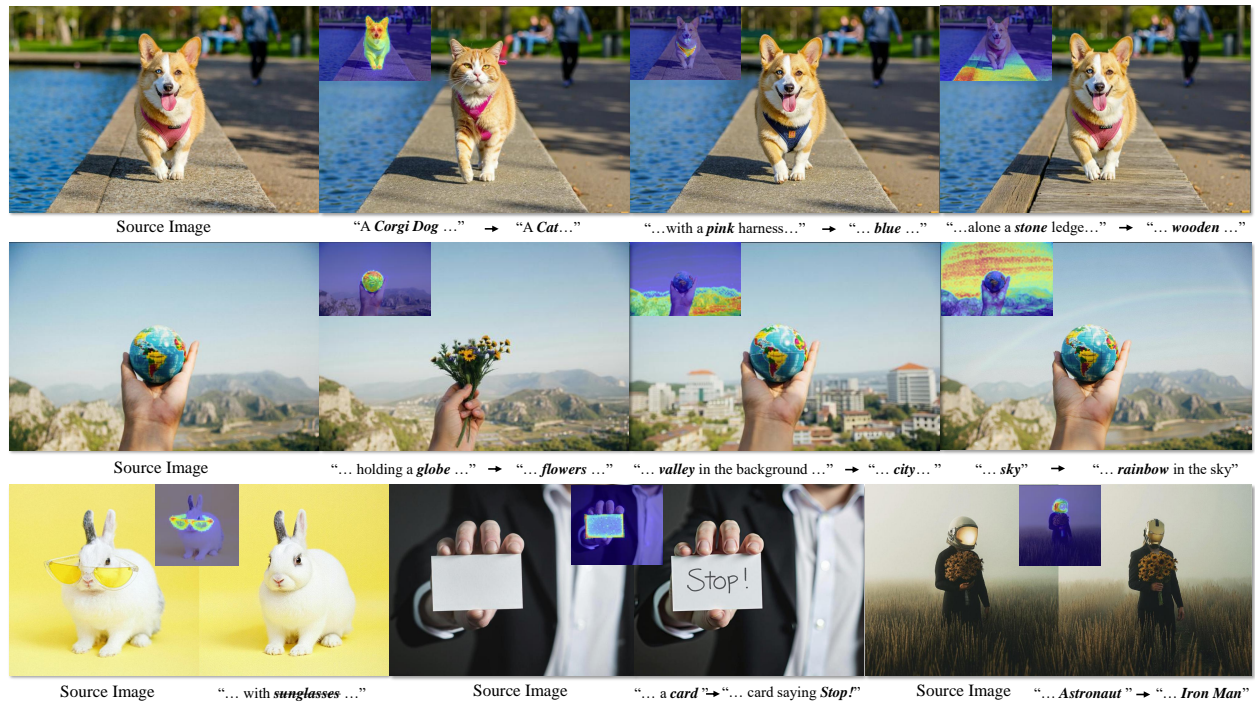

Figure 1. Given a real-world image as input, DCEdit manipulates the original content into a different semantics without additional training or tuning and could be applied to existing DiTs-based methods in a plug-and-play manner. All above results are derived from our well-designed editing benchmark RW-800.

## Abstract

*This paper presents a novel approach to improving text-guided image editing using diffusion-based models. Text-guided image editing task poses key challenge of precisely locate and edit the target semantic, and previous methods fall shorts in this aspect. Our method introduces a Precise Semantic Localization strategy that leverages visual and textual self-attention to enhance the cross-attention map, which can serve as a regional cues to improve editing performance. Then we propose a Dual-Level Control mechanism for incorporating regional cues at both feature and latent levels, offering fine-grained control for more precise edits. To fully compare our methods with other DiT-based approaches, we construct the RW-800 benchmark, featuring high resolution images, long descriptive texts, real-world images, and a new text editing task. Experimental results on the popular PIE-Bench and RW-800 benchmarks demonstrate the superior performance of our approach in preserving background and providing accurate edits.*

## 1. Introduction

In recent years, the rise of diffusion-based [15, 43–45] Text-to-Image (T2I) models [10, 22, 38, 40, 42] has driven significant progress in text-guided image editing [6, 13, 28, 33, 49]. For image editing tasks, users expect precise editing on target semantics (e.g., changing a harness from pink to blue or replacing a globe with flowers as shown in Figure 1) while maintaining other image content. This poses a key challenge: how to accurately locate and edit the target semantics in the input image.

For the T2I models, semantic localization is achieved by extracting the cross-attention map. Previous UNet-based T2I models struggled with semantic localization due to two defects: (1) weak text-image alignment, making it hard to map complex semantics accurately, and (2) limited semantic perception, effective only at low resolutions where fine details are lost. Recent Diffusion Transformer (DiT) [34] based models [10, 22] like FLUX, overcome these defects with larger text encoders [37] and multimodal DiT layers (MM-DiT) [10]. The joint self-attention module in MM-DiT enhances text-image fusion at the same scale, improving the text-image alignment. As shown in Figure 2 (a), FLUX outperforms SD-1.5 [40] and SD-XL [35] in locating fine-grained semantics.

Studies on UNet-based editing methods [6, 13, 18, 55] have proved that the cross-attention maps can be regarded as regional cues to assist the T2I model for improving editing results. However, DiT-based editing methods [3, 9, 51] overlook these cues, resulting in suboptimal performance. To address this gap, an intuitive idea is to incorporate these regional cues into the existing methods. But in practice, we find that despite its advancements, the semantic localization of FLUX still faces two critical issues: (1) incomplete activation regions, leading to poor segmentation, see Figure 2 (b), and (2) semantic entanglement, causing redundant activation, see Figure 2 (c).

To overcome these issues, we propose the Precise Semantic Localization (PSL) strategy that leverages visual and textual self-attention matrices to refine the cross-attention maps. Considering the visual self-attention matrix as the affinity between image tokens [50], we can leverage it to achieve the complement of our localized semantics extracted in the cross-attention maps. In addition, we observe that the textual self-attention matrix records the entanglement between semantics, which causes the wrong activation on the background regions. To address this entanglement, we introduce an *inverse* operation of the textual self-attention matrix to disentangle the mixed semantics for a more precise localization.

Building upon our PSL, we incorporate these refined cross-attention maps with current DiT-based editing methods [9, 51] in a plug-and-play manner. Specifically, we propose the Dual-Level Control (DLC) mechanism that introduces regional cues in both model features and diffusion latents. For feature-level control, we rescue the editing effects of current methods using the soft guidance provided by the attention map to selectively fuse model features instead of directly replacing them. For latent-level control, we implement a latent blending [1, 2] with our binarized attention map to preserve the image content in the background. As shown in Figure 1, this training and tuning free mechanism

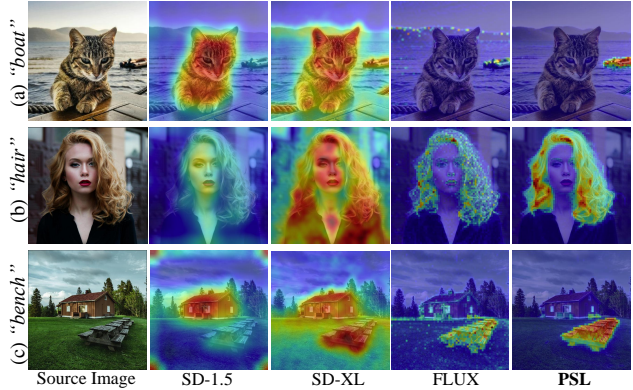

Figure 2. **Improved Semantic Localization.** (1) UNet based diffusion models such as SD-1.5 and SD-XL fail to capture detailed semantics due to limitations in the model architectures. (2) Models based on MM-DiT such as FLUX have the ability to perceive these semantics but have defects in localization. (3) Our PSL produces precise localization for semantics

achieves an accurate semantic editing based solely on text prompt without introducing extra computational costs.

To better evaluate DiT-based editing methods, we introduce the RW-800 benchmark. Comparing against the previous benchmarks [18, 33], our RW-800 demonstrates four valuable features: (1) higher resolution images ($1K$ and above *vs* $512 \times 512$), reflecting the advancement of T2I models. (2) Longer descriptive texts, enabled by DiTs' improved text encoding and text-to-image alignment. (3) Exclusively real-world images, with more complex backgrounds and richer semantic content, posing greater challenges for editing tasks. (4) Adding a new text editing task type to evaluate the ability to edit text contents.

We evaluate our DCEdit on the widely used PIE-Bench [18] and RW-800 benchmark. Evaluation results demonstrate the superior background preservation and editing performance of our method compared to previous approaches. Our contributions are:
- Precise Semantic Localization strategy for accurate semantic localization in source images,
- Plug-and-play Dual-Level Control mechanism that enhances editing using semantic localization,
- RW-800, a challenging benchmark for evaluating diffusion transformer-based editing methods.

## 2. Related works
### 2.1. Image Editing with Previous Diffusion Models

Mainstream methods [8, 13, 19, 28] based on early large-scale pre-trained Text-to-Image (T2I) diffusion models [38, 40] fulfill editing by altering the descriptive text of the source image and then regenerating it. The core challenge they face is maintaining the fidelity of the base T2I model to the edited text while ensuring consistency between the edited result and the source image. This necessitates the editing method to incorporate the source image condition. Training-based methods [5, 20, 30] involve training end-to-end models for image editing. Tuning-based methods [8, 19, 29, 54] train part of the model's parameters or specific embeddings to overfit the source image, achieving consistency during editing. Training-free methods [4, 17, 18, 32, 47, 55] are favored for their flexibility and resource-friendly attributes. They utilize diffusion mechanisms (such as inversion [43]) for the reconstruction of the input and then influence the result of editing branch by sharing information from the reconstruction branch.

### 2.2. Image Editing with Diffusion Transformers

Recently, there has been a generational upgrading in T2I models, known as Diffusion Transformers (DiTs), which employ a diffusion transformer architecture [34] and utilize a novel rectified flow [24, 26] diffusion approach. The emergence of these new generative models has spurred attempts [3, 9, 51] in editing tasks, but it has also revealed potential challenges. Some methods [12, 56] explore the attention mechanisms of diffusion transformers and specifically designed editing approaches. However, due to the lack of means to incorporate source image conditions, they perform poorly on real images. Rf-solver [51] and Fireflow [9] attempt to refine the inversion methods of rectified flow to support the reconstruction and editing of real images but with limited effect. RF-Edit proposed constructing information interaction between the inversion and sampling processes, successfully enhancing consistency with the original image at the cost of editing fidelity. Our method can provide precise semantic localization into the information interaction process, rescuing the lost editing fidelity while ensuring consistency with the original image.

### 2.3. Semantic Control for Text-driven Editing

Based on the early T2I models with UNet architecture [31], editing methods leveraged the cross-attention [13, 33, 49] maps and self-attention [6] maps generated by the model to balance the original image content with the desired editing effects. However, due to inaccurate semantic localization, these methods often excel only on simple images with prominent editing targets, and fail on real-world images with complex content. Some approaches [7, 27] introduce external masks to control editing in complex images. These methods achieve targeted area optimization by designing specific loss functions, but they significantly increase computational costs and inference time. Other methods [46, 50, 57] attempt to enhance the model's semantic localization capabilities to improve editing, but the performance gains are limited due to the constraints of the model's architecture.

## 3. Method

Given a source image $I_s$ and a pair of textual prompts $\{P_s, P_t\}$ that describe the content of the source and target images, we aim to generate an edited image that aligns with the description provided by $P_t$ while preserving the structural and semantic consistency with $I_s$. The desired semantic editing is defined by the differential word $\textit{diff}(P_s, P_t)$, whereas the remaining content should keep unchanged. To achieve this, we propose the Precise Semantic Localization strategy that extracts and refines the cross-attention maps to guide the sample process of the edited image with our Dual-Level Control mechanism. In Section 3.1 we show how PSL strategy enables refinement of the cross-attention map extracted in MM-DiT layers. In Section 3.2, we incorporate these refined maps as regional cues into both the feature space of FLUX model and the latent space in the diffusion process. Lastly, in Section 3.3, we construct a real-world benchmark dubbed *RW-800* and compare it against recent editing benchmarks.

### 3.1. Precise Semantic Localization

Recent DiTs [10, 22], such as FLUX are constructed entirely from recent advanced MM-DiT layers. FLUX incorporates joint text-image self-attention, aligns multimodal information in each MM-DiT layer. Additionally, FLUX complement the CLIP [36] text encoder with T5 [37], endowing it with significantly enhanced text comprehension capabilities. Next we introduce how to extract the text-to-image cross-attention feature maps from MM-DiT.

The MM-DiT layer employs a *joint attention mechanism* to integrate textual and visual information. First, the textual embeddings $\mathbf{T}$ and visual embeddings $\mathbf{V}$ are projected into a shared space:

$$\mathbf{Q}_T = \mathbf{T}\mathbf{W}_Q^T, \quad \mathbf{K}_T = \mathbf{T}\mathbf{W}_K^T, \quad \mathbf{V}_T = \mathbf{T}\mathbf{W}_V^T,$$

$$\mathbf{Q}_V = \mathbf{V}\mathbf{W}_Q^V, \quad \mathbf{K}_V = \mathbf{V}\mathbf{W}_K^V, \quad \mathbf{V}_V = \mathbf{V}\mathbf{W}_V^V,$$

where $\mathbf{W}_Q^T, \mathbf{W}_K^T, \mathbf{W}_V^T \in \mathbb{R}^{d_t \times d}$ and $\mathbf{W}_Q^V, \mathbf{W}_K^V, \mathbf{W}_V^V \in \mathbb{R}^{d_v \times d}$ are projection matrices, and $d$ is the shared dimension. After that, the joint attention scores $A_{joint}$ are computed by combining queries and keys from both modalities:

$$\mathbf{A}_{joint} = \text{Softmax}\left(\frac{[\mathbf{Q}_T \oplus \mathbf{Q}_V][\mathbf{K}_T \oplus \mathbf{K}_V]^\top}{\sqrt{d}}\right) \quad (1)$$

where $\oplus$ indicates the token-wise concatenation of text and visual embeddings. As shown in Figure 3, $A_{joint}$ can be split into four parts, namely textual self-attention, $V \to T$ cross-attention, $T \to V$ cross-attention and visual self-attention maps. Among them, we mainly focus on the $V \to T$ cross-attention maps (cross-attention specifically refers to this one in the following) as they most intuitively reflect the correspondence between visual and textual tokens. We first fuse each layer and head to obtain the cross-attention score map $\tilde{\mathbf{A}}^{V \to T}$:

$$\tilde{\mathbf{A}}^{V \to T} = \frac{1}{L \times H} \sum_{l \in L} \sum_{h \in H} A_{l,h}^{V \to T} \in \mathbb{R}^{M \times N} \quad (2)$$

where $A^{V \to T}$ indicates $V \to T$ cross-attention map, $L$ and $H$ are total layers and heads of the FLUX model respectively. $M$ and $N$ are the number of visual and textual tokens.

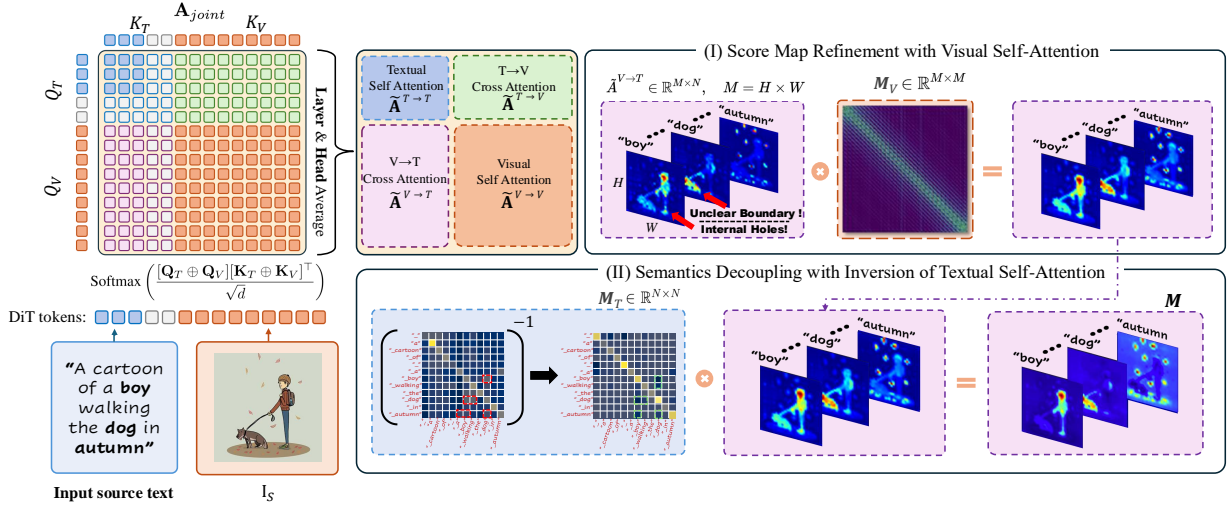

Figure 3. Illustration of PSL. The attention map $A_{joint}$ in each MM-DiT layer can be split into four parts. Among them the $\tilde{A}^{V \to T}$ can reflect the corresponding areas of semantics in the image. We further correct this attention map through visual and textual self-attention parts $\tilde{A}^{V \to V}$ and $\tilde{A}^{T \to T}$ and generate the refined attention map $\mathcal{M}$. Notably, the red anchor box on $\tilde{A}^{T \to T}$ marks the competition for attention among text tokens, which results in confusion between semantics. However, as shown by the green anchor box, the inverse of $\tilde{A}^{T \to T}$ can offset this confusion and help correct the semantic errors in the attention map.

However, as shown in Figure 3, $\tilde{\mathbf{A}}^{V-T}$ exhibits two notable shortcomings: (1) Incomplete activate regions which suffer from internal holes and lack obvious boundaries. (2) Semantic entanglement arises, leading to activations for the wrong semantics. To address these issues, we refine $\tilde{\mathbf{A}}^{V \to T}$ by leveraging the visual and textual self-attention components from $\mathbf{A}_{joint}$. We fuse the visual self-attention map as:

$$\tilde{\mathbf{A}}^{V \to V} = \frac{1}{L \times H} \sum_{l \in L} \sum_{h \in H} A_{l,h}^{V \to V} \in \mathbb{R}^{M \times M} \quad (3)$$

and reweight them to obtain a row-stochastic matrix $\mathcal{M}_V$, which captures the affinity relationships [50] among visual tokens. Besides, we note that the textual self-attention matrix records the entangle-061 ment between semantics. This entanglement of irrelevant semantics leads to redundant activation in $\tilde{\mathbf{A}}^{V \to T}$. We propose to mitigate this entanglement by leveraging the *inverse* of the fused textual self-attention matrix $\mathcal{M}_T$, where $\mathcal{M}_T \in \mathbb{R}^{N \times N}$ is obtained in the same manner as $\mathcal{M}_V$. In total, the refinement strategy of $\tilde{\mathbf{A}}^{V \to T}$ can be expressed as:

$$\mathcal{M} = norm(\mathcal{M}_V \cdot Select[(\tilde{\mathbf{A}}^{V \to T}) \cdot \mathcal{M}_T^{-1}]) \quad (4)$$

Here the $Select[\cdot]$ represents the textual token selection operation for extracting tokens associated with specific semantics, and $norm(\cdot)$ is a regular min-max normalization for map scaling.

The key idea behind multiplying $\tilde{\mathbf{A}}^{V \to T}$ with $\mathcal{M}_T^{-1}$ is $\mathcal{M}_T$ implicitly reflects the coupling relationship between semantics given in the text prompt, and thus the inversion of $\mathcal{M}_T^{-1}$ is adopted to offset this entanglement.

### 3.2. Dual-Level Control

By leveraging PSL, we obtain the refined cross-attention map $\mathcal{M}$ for specific semantics, which provides regional cues indicating where the editing effects should occur. We propose a control mechanism, dubbed Dual-Level Control, that incorporates these cues into both the *features* in FLUX model and the *latents* in diffusion process, enabling fine-grained control over the editing process.

**Inversion Process.** Image editing requires the inversion process to derive the initial noise corresponding to the source image:

$$Z_{t_{i-1}} = Z_{t_i} + (t_{i-1} - t_i) v_\theta(Z_{t_i}, t_i) \quad (5)$$

where $Z_t$ represents noise latent, timesteps $t \in \{t_K, \ldots, t_0\}$ are given in a discrete series with length $K$. $Z_{t_K} \sim \mathcal{N}(0, 1)$ are sampled Gaussian noise, and $Z_{t_0}$ is encoded latent of $I_s$. $v_\theta$ here indicates the pretrained FLUX model. PSL is applied at the beginning of inversion (generally $t_0$ only) to identify the regions associated with the replaced semantics in the source image. During the inversion process, we also store the features $V$ in the last $r$ layers and intermediate latents $Z_t$ for the following sample process.

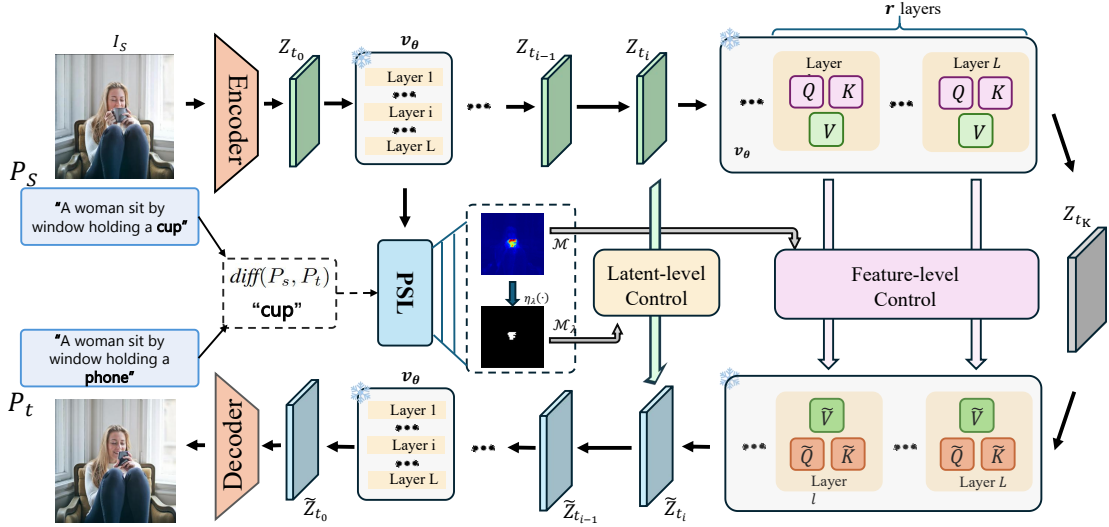

Figure 4. **The proposed Editing Pipeline.** Given the source image $I_s$ and prompt pair $\{P_s, P_t\}$, we obtain the target noise $Z_{t_K}$ through inversion process (up row). PSL works on the first inversion step incorporated with a blended word provided by $diff(P_s, P_t)$. In the sample process (down row), we guide the editing through Feature-level control and Latent-level control, which receive the $\mathcal{M}$ generated by PSL and the binary mask $M_\lambda$ respectively. Notably Feature-level control only works on last $r$ layers.

| Dataset | Numbers | Avg Res | Avg Cap | Prompt Pairs | Edit Types | All Real |
|---|---|---|---|---|---|---|
| TI2I Bench [33] | 154 | $512 \times 512$ | 5.89 | 223 | 9 | ✓ |
| ZONE [23] | 100 | $512 \times 512$ | 7.81 | 100 | 7 | ✗ |
| OIR Bench [58] | 66 | $3202 \times 3202$ | 11.66 | 200 | 5 | ✓ |
| PIE Bench [18] | 692 | $512 \times 512$ | 8.86 | 700 | 9 | ✗ |
| **RW-800 Bench** | 207 | $\mathbf{3840 \times 3690}$ | **23.14** | **800** | **10** | ✓ |

Table 1. Comparison between different public editing benchmarks. Avg Res means the raw image resolution of the source image $I_s$. Avg Cap means the average caption length of words for all images. All Real is ✗ indicates the benchmark contains synthesis images generated by generative models.

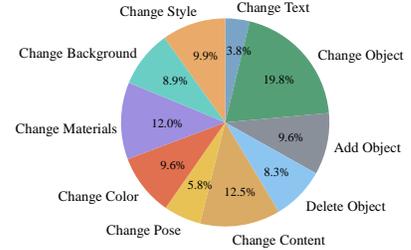

Figure 5. RW-800 Benchmark contains 10 editing categories, with an extra category of text editing. This figure shows the proportion of each category.

**Feature-Level Control with Soft Integrating.** Previous methods collect features from time steps $t \in \{t_\tau, \ldots, t_K\}$ during inversion process, and directly swap them into the sampling process to enhance consistency between the edited result and the source image, compensating for the limitations of inversion-based reconstruction. However, this approach significantly compromises editing effectiveness, as indiscriminate feature replacement discards early-stage target semantics. To address this, we introduce a unique *soft fusion* mechanism guided by refined attention map $\mathcal{M}$:

$$\hat{\mathcal{V}}_{t_i}^l = \mathcal{M} \odot \tilde{\mathcal{V}}_{t_i}^l + (1 - \mathcal{M}) \odot \mathcal{V}_{t_i}^l \quad (6)$$

As shown in Figure 4, $\mathcal{V}_{t_i}^l$ and $\tilde{\mathcal{V}}_{t_i}^l$ are projected values in $l$-th MM-DiT layer. $\mathcal{M}$ is head-wise broadcasted to fit the channel. This mechanism selectively preserves the features that are activated by the editing text during sampling, effectively avoiding the suppression of editing effects.

**Latent-Level Control for Enhanced Preservation.** In addition, considering the limitations of current rectified flow inversion [9, 41, 51] methods in reconstructing the original image, we introduce a latent-level control to further improve the image consistency. We employ a diffusion blending [1] method to fuse latents from the inversion and sampling process:

$$\hat{Z}_{t_{i-1}} = \mathcal{M}_\lambda \odot \tilde{Z}_{t_{i-1}} + (1 - \mathcal{M}_\lambda) \odot Z_{t_{i-1}} \quad (7)$$

where $\tilde{Z}_t$ is latent in sampling step $t$. We binarize $\mathcal{M}$ with $\lambda$-th percentile:

$$\mathcal{M}_\lambda = \begin{cases} 1 & if \ |\mathcal{M}| \geq \eta_\lambda(\mathcal{M}) \\ 0 & else \end{cases} \quad (8)$$

where $\eta_\lambda(\cdot)$ indicates $\lambda$-th percentage, $\lambda$ is a preset threshold. This operation is intended to prevent the background noise in $\mathcal{M}$ from affecting $Z_{t_{i-1}}$, thus preserving the consistency of the background region.

| Method | Model Structure | Structure Distance$_{\times 10^3}$ ↓ | Background Preservation | | | | CLIP Similarity | |
|---|---|---|---|---|---|---|---|---|
| | | | PSNR↑ | LPIPS$_{\times 10^3}$ ↓ | MSE$_{\times 10^4}$ ↓ | SSIM$_{\times 10^2}$ ↑ | Whole↑ | Edited↑ |
| P2P | UNet | 69.43 | 17.87 | 208.80 | 219.88 | 71.14 | 25.01 | 22.44 |
| MasaCtrl | UNet | 28.38 | 22.17 | 106.62 | 86.97 | 79.67 | 23.96 | 21.16 |
| P2P-Zero | UNet | 61.68 | 20.44 | 172.22 | 144.12 | 74.67 | 22.80 | 20.54 |
| PnP | UNet | 28.22 | 22.28 | 113.46 | 83.64 | 79.05 | 25.41 | 22.55 |
| PnP-Inv | UNet | 24.29 | 22.46 | 106.06 | 80.45 | 79.68 | 25.41 | 22.62 |
| RF-Inv | Transformer | 48.76 | 19.51 | 195.85 | 155.74 | 68.95 | 25.11 | 22.50 |
| StableFlow | Transformer | 24.95 | 21.64 | **92.28** | 115.21 | 84.94 | 24.65 | 21.70 |
| RF-Edit | Transformer | 27.70 | 23.22 | 131.18 | 75.00 | 81.44 | 25.22 | 22.40 |
| + Ours | Transformer | 26.98$_{3\%↓}$ | 24.44$_{5\%↑}$ | 114.76$_{13\%↓}$ | 59.41$_{21\%↓}$ | 83.45$_{2\%↑}$ | 25.34$_{0.4\%↑}$ | 22.67$_{1\%↑}$ |
| FireFlow | Transformer | 24.75 | 23.78 | 118.34 | 64.38 | 82.95 | 25.44 | 22.59 |
| + Ours | Transformer | **22.36**$_{10\%↓}$ | **25.41**$_{7\%↑}$ | 94.17$_{18\%↓}$ | **48.09**$_{25\%↓}$ | **85.60**$_{3\%↑}$ | **25.47**$_{0.1\%↑}$ | **22.71**$_{0.5\%↑}$ |

Table 2. Quantitative results on PIE-Bench [18]. Our approach plug-and-plays across Rf-Edit [51] and FireFlow [9], improving all metrics with the same settings. Best results are shown in bold and suboptimal results are underlined.

### 3.3. Real World Image Editing Benchmark

**Benchmark Construction.** Inspired by PIE Bench [18], we develop a novel benchmark tailored for evaluating DiT-based image editing methods. To construct a comprehensive evaluation framework, we first curate a collection of high-resolution, real-world images from available sources with proper licensing, ensuring a diverse and complex dataset suitable for testing DiTs. We then employ a state-of-the-art Vision-Language Model [16] to generate detailed source prompts $P_S$, which are refined by human annotators to correct errors and subsequently processed by an open-sourced LLM [11] to remove redundant descriptions, retaining only objective visual details. To create target prompts $P_t$, key semantic words in $P_s$ are sampled and replaced with corresponding terms from a predefined semantic library, with DeepSeek-V3 [25] ensuring linguistic fluency and coherence. For quantitative evaluation, we generate precise masks for each prompt pair $(P_s, P_t)$, where differential contents $\textit{diff}(P_s, P_t)$ are first processed using Grounded-SAM [39] to obtain initial masks, followed by an interactive segmentation pipeline for refinement. This structured approach ensures the generation of high-quality data and facilitates a rigorous evaluation of DiT-based editing methods.

**Comparison with Existing Benchmarks.** As shown in Table 1, our benchmark surpasses existing image editing datasets[18, 23, 33, 58] in several aspects. Our dataset features the largest average image size, preserving maximal visual information without cropping or downsampling. The source prompts in our dataset are significantly longer and more detailed, capturing a comprehensive semantic representation of the images. We also include the largest number of editing pairs, covering 10 distinct editing types. The distribution of these types is illustrated in Figure 5. In addition to the 9 editing types present in PIE-Bench, we introduce a new "text editing" category. This addition is motivated by the emerging capability of DiTs to accurately generate and modify text within images, which we aim to evaluate using our RW-800. More details about RW-800 are in Section A.

| Method | Structure Distance$_{\times 10^3}$ ↓ | BG Preservation | | CLIP Edited↑ |
|---|---|---|---|---|
| | | PSNR↑ | MSE$_{\times 10^4}$ ↓ | |
| RF-Inv | 28.24 | 21.46 | 92.67 | 19.90 |
| StableFlow | **15.75** | 22.04 | 95.19 | 19.83 |
| RF-Edit | 24.64 | 22.30 | 77.57 | 20.90 |
| + Ours | 22.66$_{8\%↓}$ | 23.23$_{4\%↑}$ | 62.13$_{20\%↑}$ | 20.97$_{0.3\%↑}$ |
| FireFlow | 26.59 | 21.70 | 92.26 | 20.99 |
| + Ours | 21.07$_{21\%↓}$ | **23.59**$_{9\%↑}$ | **57.50**$_{38\%↑}$ | **21.14**$_{1\%↑}$ |

Table 3. Comparison with DiT-based editing methods on RW-800. Best results are shown in bold, suboptimal results are underlined.

## 4. Experiments

In Section 4.1, we evaluate the editing capabilities of the method on widely-used editing benchmarks as well as on our RW-800. In Section 4.2, we quantitatively compare the semantic localization capabilities of PSL with methods based on UNet-based models. In Section 4.3 we discuss the effectiveness of each components in our editing pipeline. The implementation details are in Section B.

### 4.1. Comparison for Image Editing

**Quantitative Comparison on PIE-Benchmark.** To comprehensively evaluate the performance of our proposed method, we first conduct experiments on the widely adopted

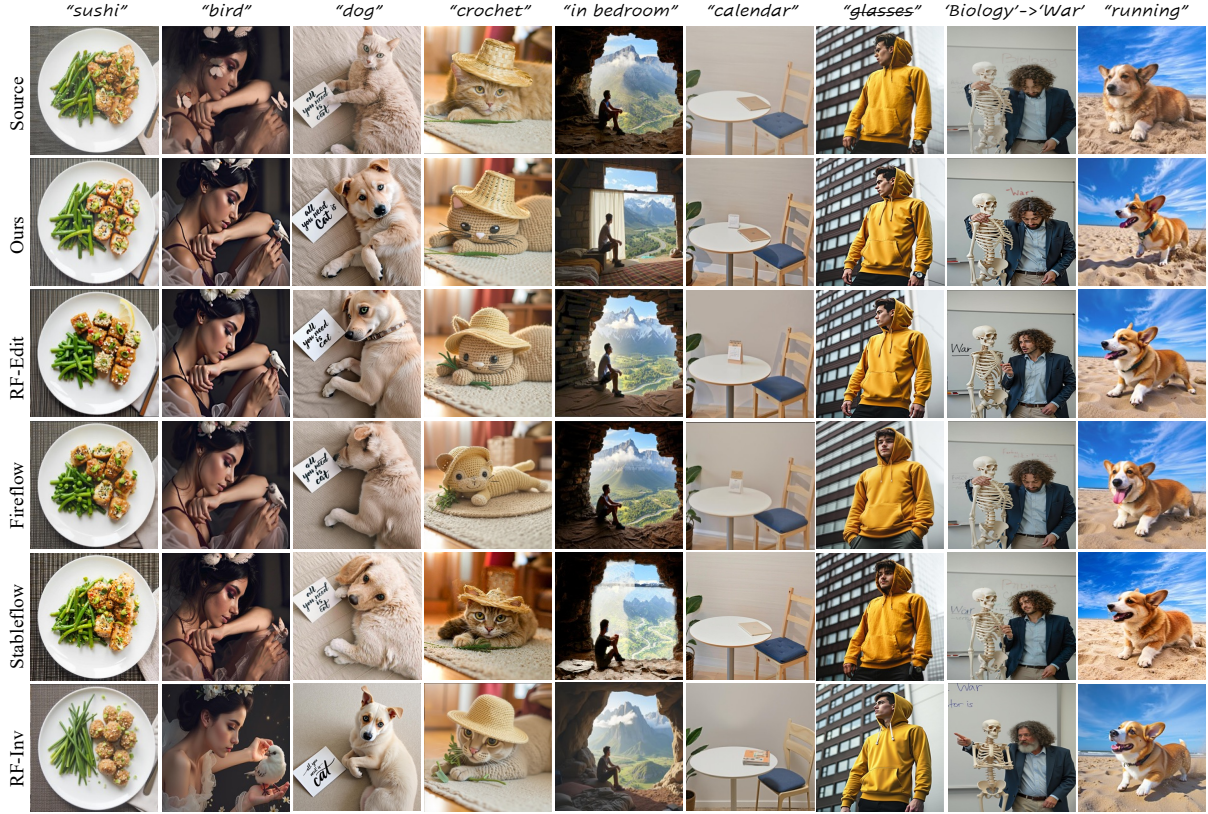

Figure 6. Comparison between editing methods in our RW-800 dataset. Please zoom in for a better view.

| Method | PIE | | RW-800 | |
|---|---|---|---|---|
| | MSE$_{\times 10^2}$ | IoU(%) | MSE$_{\times 10^2}$ | IoU(%) |
| SD1.5 Cross-Attn. | 24.70 | 16.12 | 15.74 | 13.87 |
| SDXL Cross-Attn. | 19.57 | 21.82 | 13.48 | 24.80 |
| FLUX Cross-Attn. | 24.05 | 39.48 | 10.28 | 37.99 |
| + VSA | 18.97 | 41.02 | 7.42 | 53.83 |
| + TSA (**PSL**) | 17.84 | 41.28 | 6.91 | 56.01 |

Table 4. Comparison of semantic localization capabilities. We conduct a comparative analysis of the semantic localization capabilities of native cross-attention maps across different text-to-image models, and perform ablation studies to investigate the contributions of components in PSL.

PIE-Bench [18]. For comparison, we select a range of baseline methods, including classical training-free editing approaches based on Diffusion UNet, such as P2P [13], MasaCtrl [6], P2P-zero [33], PnP [49], and the improved DDIM inversion scheme PnP-Inv [18]. Additionally, we compare our method with recent DiT-based editing techniques, including RF-Inv [41], Stable Flow [3], RF-Edit [51], and Fireflow [9]. Results are shown in Table 2. Notably, our approach operates in a plug-and-play manner on RF-Edit and Fireflow, simultaneously enhancing background consistency and editing quality without introducing additional computational overhead. This demonstrates the versatility and efficiency of our method in improving existing state-of-the-art frameworks.

**Quantitative Comparison on RW-800 Benchmark.** To further validate the robustness and generalization capability of our method, we conduct an extensive evaluation on the challenging RW-800 benchmark, comparing it with state-of-the-art DiT-based editing approaches [3, 9, 41, 51]. Experimental results demonstrate that our method significantly enhances the performance of both RF-Edit and Fireflow. Specifically, in Table 3, our approach achieves remarkable improvements in structural similarity [48], reducing the background MSE of RF-Edit and Fireflow by 20% and 38%, respectively. Moreover, it simultaneously enhances the editability of target regions, maintaining a balanced improvement across multiple evaluation metrics. Stable Flow achieves content preservation with the original image through attention injection in vital layers, resulting in closer structural distance and better SSIM scores compared to the source image. However, even with a limited number of vital layers, this strong control mechanism significantly compromises its editing capability, reflects on low CLIP score.

**Qualitative Comparison.** We qualitatively compare with other DiT-based editing methods on the RW-800 benchmark. As shown in the Figure 6, RF-inv's editing will bring a large difference to the original image, while Stable Flow's editing effect is not significant. Our method has a more obvious editing effect than RF-Edit and Fireflow, and maintains the original image in the background area. More qualitative comparison on PIE-Bench are in Section C

| Baselines | | + *Feature-level control* | | | + *Latent-level control* | | Structure | CLIP |
|---|---|---|---|---|---|---|---|---|
| StableFlow | FireFlow | w\$\mathcal{M}_\lambda$ | w\$GT\ mask$ | w\$\mathcal{M}$ | 2-step | 5-step | Distance$_{\times 10^3}$ ↓ | Edited↑ |
| ✓ | | | | | | | 15.75 | 19.83 |
| | ✓ | | | | | | 26.59 | 20.99 |
| | ✓ | ✓ | | | | | 29.30 | 21.31 |
| | ✓ | | ✓ | | | | 29.14 | 21.37 |
| | ✓ | | | ✓ | | | 26.74 | 21.28 |
| | ✓ | | | ✓ | ✓ | | 21.07 | 21.14 |
| | ✓ | | | ✓ | | ✓ | 11.1 | 20.53 |

Table 5. Ablation studies on the role of Feature-level control and Latent-level control. $GT\ mask$ is provided by RW-800 Benchmark. We claim the effectiveness of using the model inner attention map in Feature-level control.

### 4.2. Evaluation of Semantics Localization

**Settings.** To evaluate the semantic localization capability of PSL, we conduct experiments on two editing benchmarks: PIE-Bench [18] and RW-800 Benchmark. Both benchmarks provide paired image-text data along with manually annotated masks for editing regions, enabling comprehensive assessment of background preservation and foreground editing performance. Leveraging these masks, we quantitatively analyze the model's performance by computing the Mean Square Error (MSE) between the attention maps and the ground truth masks, as well as the Intersection-over-Union (IoU) score after binarizing these attention maps.

**Quantitative Comparison on Editing Benchmarks.** For comparison, we select diffusion UNet based text-to-image diffusion models as baselines, including SD-1.5 [40] and SD-XL [35], both of which allow extraction of attention maps from their cross-attention layers. Furthermore, we systematically compare the performance of directly utilizing Flux's joint self-attention mechanism [10, 22] with our refinements incorporating visual self-attention and textual self-attention parts, respectively. The $1^{st} - 3^{rd}$ rows of the Table 4 demonstrate that FLUX, based on the MM-DiT architecture, significantly outperforms both SD-1.5 and SD-XL built on UNet in terms of semantic localization, achieving notably higher Intersection-over-Union (IoU) scores. This improvement highlights the superior capability of FLUX in aligning semantic information with visual content. Furthermore, the integration of visual self-attention and textual self-attention components into FLUX's cross-attention mechanism leads to a substantial enhancement in localization accuracy. These results underscore the effectiveness of our proposed architectural modifications in achieving precise and robust semantic localization, which is crucial for high-quality image editing tasks.

### 4.3. Ablation Studies and Analysis.

We conduct ablation studies to evaluate the impact of various components on the model's editing performance using real images. All experiments are performed on the RW-800 benchmark, based on the 8-step Fireflow method.

As shown in the $1^{st} - 3^{rd}$ rows in Table 5, in the case of feature-level control only, we test guiding the model with PSL's binary mask, which could improve editing but also increase structure distance, likely due to segmentation inaccuracies. Using the ground-truth masks from the benchmark couldn't result in significant improvements. In contrast, employing the score map $\mathcal{M}$ for control reduces structure distance while maintaining high editing capability. This improvement is attributed to the limitations of binary masks, which disrupt feature representation during mixing, causing deviations. The soft fusion approach with $\mathcal{M}$ preserves feature integrity, ensuring consistent and high-quality edits. Additionally, the continuous map provides richer information and more precise guidance for the editing process compared to binary masks.

Then, we further integrate the latent-level control mechanism into the framework. As shown in the $4^{th} - 5^{th}$ rows, we apply the binarized mask $\mathcal{M}_\lambda$ for latent blending during the first two sampling steps, preserving background information by bypassing early denoising stages. This mechanism significantly improves consistency between the edited and original images with minimal loss in editability. Extending the control to more sampling steps further enhances consistency. When applied for 5 steps, our method achieves a lower structural distance than StableFlow, while outperforming it by nearly 4% in CLIP score. These results highlight the effectiveness of the latent-level control strategy in balancing editing quality and image consistency.

### 5. Conclusion

This paper introduces a novel DCEdit tailored for text-guided image editing. Through the proposed Precise Semantic Localization strategy, we enhance the quality of the extracted cross-attention maps, which become precise regional cues to assist image editing. Our Dual-Level Con-trol mechanism effectively incorporates the regional cues at both feature and latent levels, boosting the performance of DiT-based editing methods. Additionally, the construction of the RW-800 benchmark provides a comprehensive evaluation tool that challenges existing methods and highlights the superiority of our approach in real-world scenarios. Our results demonstrate improvements over previous methods in terms of both background preservation and editing quality by a large margin, making DCEdit a promising solution for the future of text-to-image editing.

# DCEdit: Dual-Level Controlled Image Editing via Precisely Localized Semantics

## Supplementary Material

## A. Details about RW-800 Benchmark

### A.1. Motivation

The ability of Diffusion Transformers (DiTs) [34] to process longer prompts, which inherently contain richer information, necessitates the use of higher-resolution images as inputs. Existing benchmarks [18, 23, 33, 58] fall short in this regard, as they predominantly feature low-resolution images paired with brief captions, making them unsuitable for evaluating the full potential of DiTs. Furthermore, higher-resolution images encompass more diverse content, enabling a wider range of editing possibilities. This motivates the need for a benchmark that supports multiple editing directions for a single image, thereby facilitating a comprehensive assessment of DiTs' capabilities. To fill this gap we proposed RW-800 Benchmark.

### A.2. Procedure of Benchmark Construction.

**Source Image Obtaining.** To address the limitations of existing datasets, we curated a collection of high-resolution, real-world images from open copyright materials webset Pexels at: https://www.pexels.com/ with proper licensing. These images were carefully selected to ensure a high degree of complexity and diversity, providing a challenging testbed for DiTs.

**Caption Generation and Cleaning.** We employed CogVLM2 [16] to generate the detail caption $P_S$ for source image. The input prompt of the VLM is *'In one sentence, describe content that is visible in the image. Prioritize accuracy. Start with "the image shows"'* and then take out *"the image shows"*. These captions are refined subsequently by human annotators to supplement the missing visual content and correct false description. Then we use GLM-4-9b [11] model to remove redundant descriptions. We define this redundant removing task with the following prompt:

> Task: Remove redundant information, subjective descriptions, and unnecessary details from the following image caption to make it more concise and objective. Focus on keeping only the essential information about the image.
>
> Example: Original Caption: "The beautiful and stunning sunset with a red sky is very nice, with lots of clouds and a few birds flying in the sky." Cleaned Caption: "A sunset with a red sky and a few birds in the sky."
>
> Original Caption: "There is a large, blue, shiny car on the street. The car looks really amazing and it is very noticeable." Cleaned Caption: "A blue shiny car on the street."
>
> Input Caption: [Insert the image caption here] Output Caption:

**Target Prompt Generation.** To create target prompts $P_t$, we sampled and replaced key semantic words in $P_s$ with corresponding terms from a predefined semantic library. We adopted the strategy in HeadRouter [56] to build the semantic library, but our goal was to build an image editing benchmark to evaluate the editing quality. The modified prompts were then processed by DeepSeek-V3 [25] to ensure linguistic fluency and coherence. The prompt we use here is:

> Task: Optimize sentence 2 while maintaining the same structure as sentence 1. Ensure that sentence 2 becomes grammatically correct and fluent without altering the meaning of sentence 1.
>
> Example: Sentence 1: "The cat jumped over the wall." Sentence 2: "The cat stand over the wall." Optimized Sentence 2: "The cat stand on the wall."
>
> Input Sentence 1: [Insert sentence 1 here] Input Sentence 2: [Insert sentence 2 here] Output Sentence 2:

**Mask Generation for Evaluation.** For each prompt pair $\{P_s, P_t\}$, we generated precise masks to facilitate quantitative evaluation. Specifically, the differential contents $\textit{diff}(P_s, P_t)$, are first processed using Grounded-SAM [21, 39] to obtain initial masks. This mask are further manually refined through an online interactive segmentation pipeline Roboflow: https://roboflow.com/ to ensure the correctness of annotation.

## B. Implementation Details

**Third-party Implementations of Baselines.** As explained in Section 4.2 of the main paper, we compare our method against the following baselines. Diffusion UNet based methods include: DDIM [43] + P2P [13], DDIM + MasaCtrl [6], DDIM + PnP [49], DDIM + P2P-zero [33], PnP-Inv [18], and Diffusion Transformer based methods include: RF-Inv [41], Stable Flow [3], RF-Edit [51], Fireflow [9].

We reference the following third-party implementations in this project:

- **Diffusion UNet based methods.** Here we use systematically reproduced version by: https://github.com/cure-lab/PnPInversion, which is the official implementation of PIE-Bench.

| Editing Type | Source Image | Source Prompt | Target Prompt | Blending Word | Editing Mask | Editing Type | Source Image | Source Prompt | Target Prompt | Blending Word | Editing Mask |
|---|---|---|---|---|---|---|---|---|---|---|---|
| 1 Change Object | 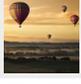 | Multiple [hot air balloons] floating above a mist-covered countryside…over the landscape. | Multiple [space crafts] floating above a mist-covered countryside…over the landscape. | hot air balloons _space crafts | 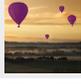 | 6 Change Color | 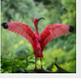 | A vividly colored [red] bird with outspread wings, standing on a branch amidst a lush green forest backdrop. | A vividly colored [pink] bird with outspread wings, standing on a branch amidst a lush green forest backdrop. | bird_bird | 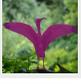 |
| 2 Add Object | 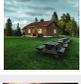 | A brown wooden house with a red roof, surrounded by…with wooden picnic benches on the lawn. | A brown wooden house with a red roof, surrounded by…with wooden picnic benches and [kids on the lawn]. | lawn_lawn | 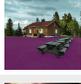 | 7 Change Materials | 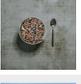 | A bowl filled with cigarette butts placed on a textured surface, accompanied by a spoon. | A bowl filled with cigarette butts placed on a textured surface, accompanied by a [wooden] spoon. | spoon_spoon | 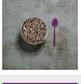 |
| 3 Delete Object | 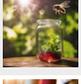 | A [bee hovering] near a glass jar on a wooden surface…against a backdrop of green foliage illuminated by sunlight. | A glass jar on a wooden surface…against a backdrop of green foliage illuminated by sunlight. | bee_[null] | 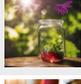 | 8 Change Background | 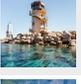 | A multi-level observation tower…against a backdrop of clear [blue skies] and turquoise waters… | A multi-level observation tower…against a backdrop of clear [dark skies] and turquoise waters… | skies_skies | 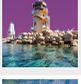 |
| 4 Change Content | 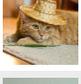 | An orange cat lying on a beige rug, wearing a [woven straw hat]…with a wooden floor and a red curtain. | An orange cat lying on a beige rug, wearing a [top hat]…with a wooden floor and a red curtain. | hat_hat | 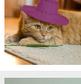 | 9 Change Style | 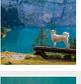 | A turquoise lake nestled between rugged mountains, a waterfall cascading… | [A kid's drawing of] a turquoise lake nestled between rugged mountains, a waterfall cascading… | [null] | 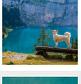 |
| 5 Change Pose | 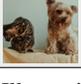 | A Yorkshire Terrier dog sitting on a beige fabric with its [tongue out]…against a pale blue wall. | A Yorkshire Terrier dog sitting on a beige fabric with its [mouth closed]…against a pale blue wall. | tongue out _mouth | 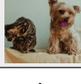 | 10 Change Text | 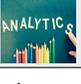 | The word [ANALYTICS] spelled out…and a hand pointing towards… | The word [AGI] spelled out… and a hand pointing towards… | ANALYTICS _AGI | 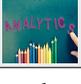 |

Figure 1. Illustration of our RW-800 dataset. Our dataset consists of 10 categories, covering most editing tasks. The blending word and editing mask indicate the objects to be edited. Some tasks, such as "Delete object" and "Change style," may not include the above two categories due to the lack of a specific object to edit. The prompts in the images have been partially omitted for display purposes.

- **RF-Inv** officially implementated at Diffusers: https://github.com/huggingface/diffusers.
- **Stable Flow** officially implemented at: https://github.com/snap-research/stable-flow.
- **RF-Edit** officially implemented at: https://github.com/wangjiangshan0725/RF-Solver-Edit.
- **Fireflow** officially implemented at: https://github.com/HolmesShuan/FireFlow-Fast-Inversion-of-Rectified-Flow-for-Image-Semantic-Editing.

**Evaluation Metrics.** To illustrate the effectiveness and efficiency of our DLC, we follow the PIE-Bench and use seven metrics covering three aspects: structure distance [49], background preservation (PSNR, LPIPS [59], MSE, and SSIM [52] outside the annotated editing mask), edit prompt-image consistency (CLIPSIM [53] of the whole image and regions in the editing mask). For simplicity, we omitted some indicators with repeated metrics in the RW-800 experiment and ablation study.

**Method Implementation Details.** Our plug-and-play method works on RF-Edit and Fireflow, and the same step settings and classifier-free guidance (cfg) [14] are used for comparison. In the experimental settings, we follow the recommended settings, with the inversion and sample steps of Fireflow being 8, and the inversion and sample steps of RF-Edit being 15. In the experiments on PIE, we set the cfg value to 2 according to the recommendation of Fireflow, and the cfg value of RF-Edit to 3. On RW-800, in order to make the editing effect more significant, we uniformly set the cfg value to 3. Our method performs 1-step Feature-level Control and 3-step Latent-level Control during sampling on Fireflow, and 3-step Feature-level Control and 3-step Latent-level Control on RF-Edit.

## C. More Qualitative Comparison

**Comparison on PIE-Bench.** Here we supplement the qualitative comparison on PIE-Bench in Fig 2. Compared with the diffusion UNet-based method, transformer-based editing methods such as Stable Flow, RF-Edit and Fireflow have more comprehensive effect improvements. The feature-level control in our DCL can effectively release the editing effect suppressed in value injection, while the latent-level control can specifically retain the information in the background, thereby achieving a better balance between editing effect and background consistency.

**Visualization Results of PSL.** In order to more fully demonstrate the effectiveness of PSL, here we provide more visualization effects to supplement the ablation study. As shown in Figure 3, After the refinement of visual self-attention, the corresponding areas of each semantic in the cross-attention map can be completed, but at the same time, the wrong area will be activated, resulting in the wrong semantic localization. After adding textual self-attention, various semantics are decoupled, further improving the accuracy of semantic localization.

## D. Limitations and Future Works

DLC will converge back to the original editing method in some tasks that modify the entire image (such as image style transfer). This is because there is no corresponding semantics in the original image, so it it unattainable for the correspondent attention map. In future work, we can try to extract and optimize the distribution information of model features to complete image stylization tasks.

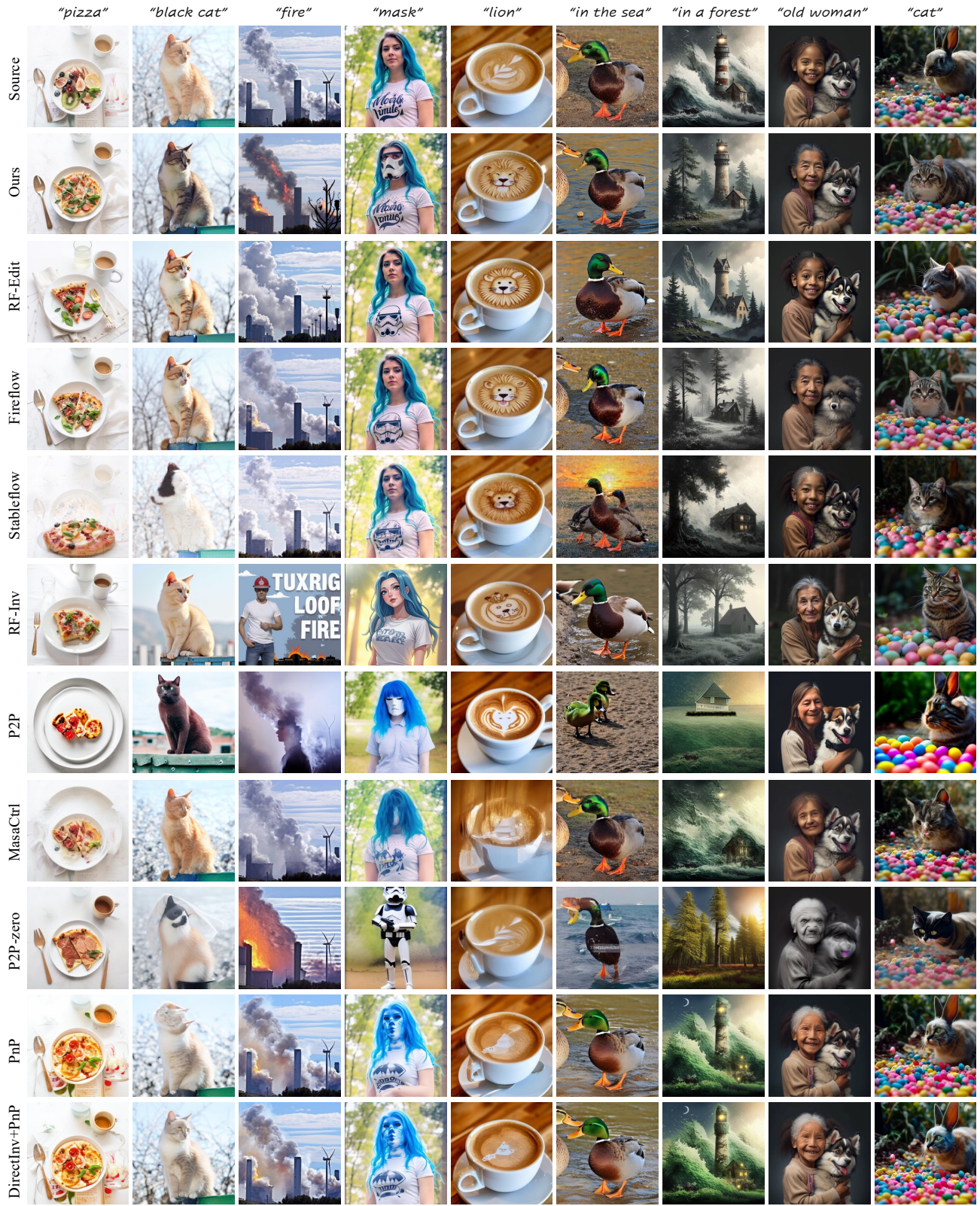

Figure 2. Qualitative comparison on PIE-Bench with both Diffusion UNet-based methods and DiT-based methods.

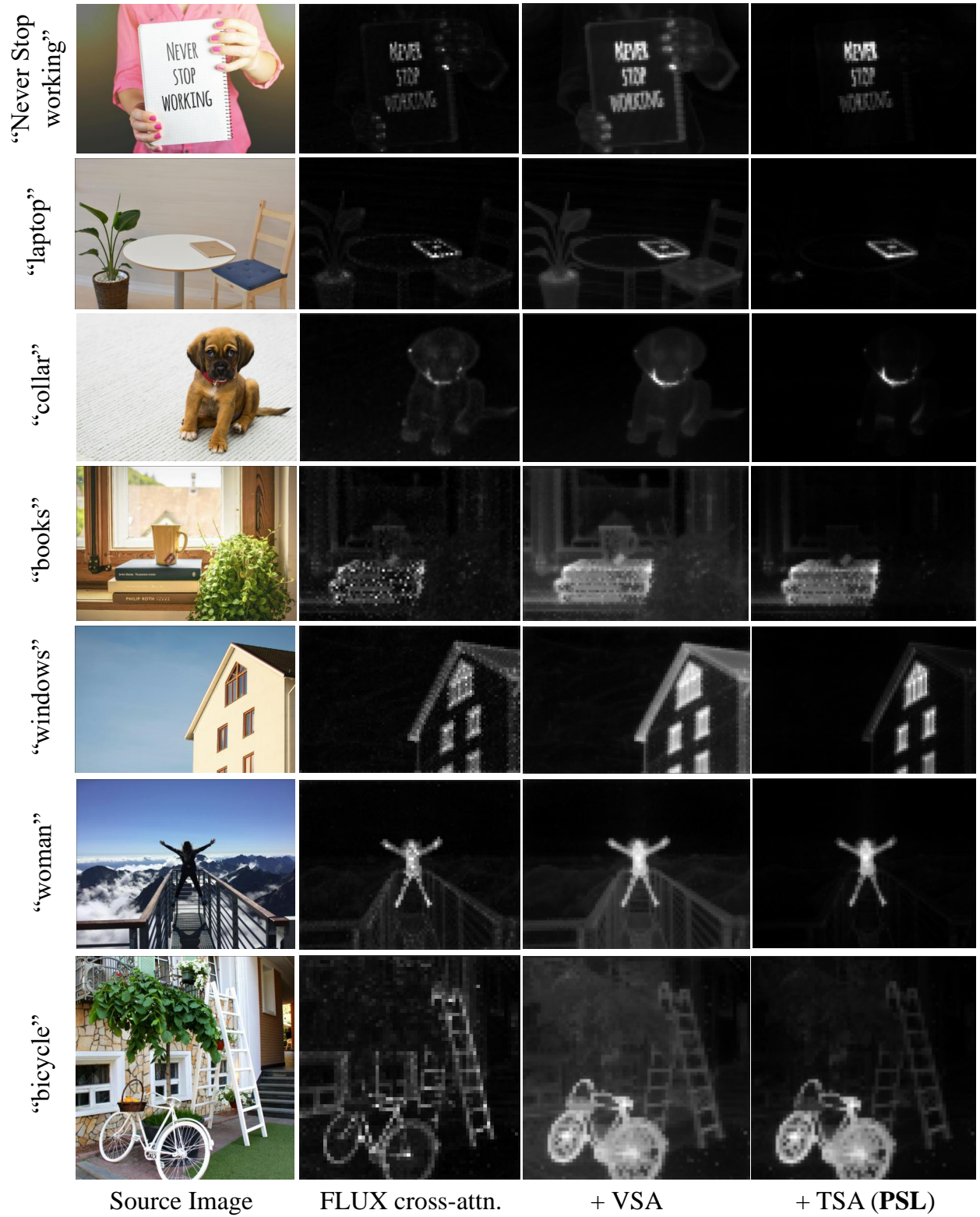

Figure 3. **Ablations Qualitative Comparison of PSL.** Through refinement of visual self-attention and textual self-attention, PSL improves the quality of cross-attention map naively generated by MM-DiT layers of FLUX. Each attention map is activated by the blended word in the left. VSA and TSA indicates using visual self-attention and text self-attention respectively. All shown cases are in our RW-800